\documentclass{article}

\usepackage{arxiv}

\usepackage[utf8]{inputenc} 
\usepackage[T1]{fontenc}    
\usepackage{hyperref}       
\usepackage{url}            
\usepackage{booktabs}       
\usepackage{amsfonts}       
\usepackage{nicefrac}       
\usepackage{microtype}      
\usepackage{amssymb}
\usepackage{amsmath}
\usepackage{amsthm}
\usepackage{graphicx}
\usepackage{color}
\usepackage{xcolor}                   
\usepackage{framed}
\usepackage{mdframed}
\usepackage{enumitem}
\usepackage{xspace}

\newcommand{\smoothgrad}{{\sc SmoothGrad}\xspace}

\mdfdefinestyle{MyFrame}{%
  align=center,
  userdefinedwidth=.8\textwidth,
  skipabove=10pt,
  skipbelow=10pt,
  innertopmargin=4pt,
  innerbottommargin=4pt,
  innerrightmargin=4pt,
  innerleftmargin=4pt,
  leftmargin = 4pt,
  rightmargin = 4pt,
  backgroundcolor=gray!10!white,
  linecolor=red!60!black,
  linewidth=2pt,
  rightline=false,
  leftline=false
}

\title{Towards Auditability for Fairness in Deep Learning}

\author{%
  Ivoline C. Ngong \\
  Department of Computer Engineering\\
  Konya Technical University\\
  \texttt{ivolinengong@gmail.com} \\
  \And
  Krystal Maughan \\
  Department of Computer Science\\
  University of Vermont\\
  \texttt{Krystal.Maughan@uvm.edu} \\
  \And
  Joseph P. Near \\
  Department of Computer Science\\
  University of Vermont\\
  \texttt{jnear@uvm.edu} \\
}

\begin{document}

\maketitle

\begin{abstract}
  Group fairness metrics can detect when a deep learning model behaves
  differently for advantaged and disadvantaged groups, but even models
  that score well on these metrics can make blatantly unfair
  predictions. We present \emph{smooth prediction sensitivity}, an
  efficiently computed measure of individual fairness for deep
  learning models that is inspired by ideas from interpretability in
  deep learning. smooth prediction sensitivity allows individual
  predictions to be audited for fairness. We present preliminary
  experimental results suggesting that smooth prediction sensitivity
  can help distinguish between fair and unfair predictions, and that
  it may be helpful in detecting blatantly unfair predictions from
  ``group-fair'' models.
\end{abstract}

\section{Introduction}

It is known that systems today experience algorithmic bias that is tied to the data collection process, the data modelling process, and historical or systemic contexts of bias. As systems become more automated in practice, legal definitions of algorithmic discrimination will require systems to be audited for bias and fairness to determine whether they are fair or unfair in their predictions. Robustness in the evaluation of algorithms will be integral for the safety and fairness of populations and in maintaining systems that serve all people.

A key capability is the ability to \emph{audit} individual decisions made by an algorithm---to identify whether a potentially unfair decision was made based specifically on membership in a particular group (for example, if a person belonged to a particular race, gender or other protected class), or because of a property of the individual.

Consider an individual who has, through an automated system, not been admitted to a particular school. In a legal claim, they have identified that an automated algorithm has unfairly discriminated against them in a manner that they deemed to be unfair. In an ideal auditory system, their challenge of such a decision would involve determining whether \emph{that recommendation} of non-admittance was dependent on any group attribution bias (as an individual in a specific group to which they belong) or as an individual.


The majority of work on fairness in AI has focused on ensuring
\emph{group fairness}~\cite{calders2009building,
  woodworth2017learning, zafar2017fairness, agarwal2018reductions,
  russell2017worlds, celis2019classification, beutel2017data,
  shankar2017no, zhang2018mitigating, wadsworth2018achieving,
  zemel2013learning, louizos2015variational, lum2016statistical,
  calmon2017optimized, feldman2015computational,
  hardt2016equality}. However, group fairness does not imply the
ability to audit \emph{specific} decisions made by AI-based systems.
As Lipton et al.~\cite{LiptonMC18} demonstrate, {even ``fair'' models
  can make \emph{unfair decisions} for some individuals}---even when
they score well on metrics for group fairness. Group fairness---both
metrics and mitigation---are therefore insufficient to provide the
kind of auditability described above.


Maughan and Near previously developed \emph{prediction sensitivity}~\cite{DBLP:journals/corr/abs-2009-13650}, a measure of individual fairness based on computing partial derivatives of individual predictions. In this work, we make two novel contributions building on prediction sensitivity. First, we propose \emph{smooth prediction sensitivity}, an extension to prediction sensitivity that draws on ideas from interpretability~\cite{morch1995visualization, baehrens2010explain, simonyan2013deep, sundararajan2017axiomatic, smilkov2017smoothgrad, hooker2019benchmark} to improve our measure of fairness when the prediction's gradient is not already smooth.

Second, we conduct a larger empirical evaluation of our technique, using multiple datasets and model architectures. The results suggest that prediction sensitivity (both smoothed and unsmoothed) is useful for distinguishing between fair and unfair predictions; our results also suggest that our current formulation of prediction sensitivity is more effective for simpler architectures (e.g. linear models) than it is for more complex ones (e.g. convolutional networks).

In addition, we evaluate smooth prediction sensitivity on ``fair'' models---models trained \emph{with} an existing bias mitigation technique, that score well on group fairness metrics. Our results suggest that (1) these ``fair'' models indeed make some \emph{unfair} predictions, and (2) smooth prediction sensitivity may be a useful tool to help detect them.

\section{Background \& Related Work}

\paragraph{Deep learning.}
In this paper, we focus on machine learning models represented by \emph{artificial neural networks}~\cite{goodfellow2016deep}. A model $\mathcal{F}$ is parameterized by a set of \emph{weights} $\theta$ which are optimized during training; we write $\mathcal{F}(\theta, x)$ to represent a \emph{prediction} made by the trained model on an example $x$. Deep learning models are typically trained by optimizing a \emph{loss function} $\mathcal{L}$ using a ground-truth label $y$ for each example. We write the loss for an example $x,y$ as $\mathcal{L}(\mathcal{F}(\theta, x), y)$. During training, the loss is used to update the weights $\theta$ so that loss is reduced during the next training epoch.

\paragraph{Automatic differentiation.}
\emph{Automatic differentiation}~\cite{baydin2017automatic} is a computational method used to evaluate the derivative of a function efficiently. Automatic differentiation is normally used for computing the \emph{gradient} of the loss with respect to the model's weights: $\nabla_\theta(\mathcal{L}(\mathcal{F}(\theta, x), y))$.
%
%
This gradient is a vector containing the partial derivative of the model's output with respect to each of the weights. Automatic differentiation systems in modern deep learning frameworks are specifically designed to efficiently compute gradients for functions with many inputs, and they are usually used to calculate gradients during training.

\paragraph{Fairness in machine learning.}
The bulk of previous work on fairness in machine learning attempts to improve \emph{group fairness} at training time, often by the introduction of new kinds of regularization~\cite{calders2009building, woodworth2017learning, zafar2015fairness, zafar2017fairness, agarwal2018reductions, russell2017worlds, celis2019classification, beutel2017data, shankar2017no, zhang2018mitigating, wadsworth2018achieving, celis2019improved, zemel2013learning, louizos2015variational, lum2016statistical, adler2018auditing, calmon2017optimized, feldman2015computational, hardt2016equality}. Many of these approaches are suitable for deep learning, and have been empirically validated using the metrics described above.

Existing approaches focus on notions of group fairness, and are validated using metrics for group fairness. As a result, they can sometimes produce models that give blatantly \emph{unfair} predictions for specific individuals, even though they score well on group fairness metrics~\cite{dwork2012fairness, LiptonMC18}.

\paragraph{Interpretability in deep learning.}
The related field of \emph{interpretable AI} has a similar goal: to enable auditability by understanding \emph{why} a model made a specific prediction. The problem of interpretability has been studied extensively in the setting of image classification, where the goal is to understand which pixels of the image were most important in deciding its class. Numerous gradient-based approaches have been proposed~\cite{morch1995visualization, baehrens2010explain, simonyan2013deep, sundararajan2017axiomatic, hooker2019benchmark}; our work draws on the ideas in \smoothgrad~\cite{smilkov2017smoothgrad} (details in Section~\ref{sec:smooth-pred-sens}).



\section{Prediction Sensitivity}

\begin{mdframed}[style=MyFrame]
\begin{center}
  \textbf{Hypothesis:} Predictions that rely heavily on the protected
  attribute will tend to disadvantage the protected group.
\end{center}
\end{mdframed}

As an initial step towards measuring individual fairness in deep learning models, Maughan and Near previously proposed \emph{prediction sensitivity}~\cite{DBLP:journals/corr/abs-2009-13650}, which attempts to quantify the \emph{extent to which a single prediction depends on the protected attribute}. We hypothesize that models which rely heavily on the value of the protected attribute are likely to make different predictions for members and non-members of the advantaged group, and that prediction sensitivity may therefore be a useful measure of individual fairness. Prediction sensitivity is related to the individual metric proposed by Dwork et al.~\cite{dwork2012fairness}, but prediction sensitivity can be efficiently computed for artificial neural networks. This section gives a brief summary of prediction sensitivity.


\paragraph{Formal Definition.}
Formally, we assume the existence of a neural network architecture $\mathcal{F}$ such that for a vector of trained weights $\theta$ and feature vector $x$, we can make a prediction $\hat{y}$ as follows: $\hat{y} = \mathcal{F}(\theta, x)$.
%
%
For such a model $\mathcal{F}$, we define the prediction sensitivity with respect to attribute $a \in x$ as the partial derivative:
\[ \Big\lvert \frac{\partial}{\partial a} \mathcal{F}(\theta, x) \Big\rvert \]

\paragraph{Computing Prediction Sensitivity.}
The same automatic differentiation libraries commonly used to compute gradients of the loss during training can also be used to efficiently compute prediction sensitivity. Given a loss function $\mathcal{L}$ and a training example $x, y$, for current weights $\theta$, the training process might compute the gradient: $\nabla_\theta(\mathcal{L}(\mathcal{F}(\theta, x), y))$.
%
%
Here, we write $\nabla_\theta$ to denote the gradient with respect to each weight in $\theta$. Thus the gradient contains partial derivatives of the \emph{loss} with respect to the \emph{weights}.
To compute prediction sensitivity for a feature vector $x$, we want to obtain the partial derivative of the \emph{prediction} with respect to \emph{one feature}. Given a trained model consisting of $\mathcal{F}$ and $\theta$, we can compute prediction sensitivity for the feature $a$ as: $\lvert \nabla_a (\mathcal{F}(\theta, x)) \rvert$.
%
%
This value can be efficiently computed using existing deep learning frameworks.




\section{Smooth Prediction Sensitivity}
\label{sec:smooth-pred-sens}

Prediction sensitivity is intended to measure how sensitive a prediction is to the value of the protected attribute. We hypothesize that relying heavily on the protected attribute in making a prediction may indicate that the prediction \emph{would have been different} if the individual had been a member of a different group. 

However, prediction sensitivity only measures this value at \emph{one particular point}---the example whose label is being predicted---and small changes to one or more features could potentially increase or decrease prediction sensitivity significantly (in fact, the amount of this change is potentially unbounded). This effect could result in a failure of prediction sensitivity to align well with the counterfactual statement above: prediction sensitivity could be low at the example itself, but very high in its neighborhood. In order to be useful for distinguishing between fair and unfair predictions, we must assume that prediction sensitivity is fairly \emph{smooth}---i.e. that small changes in features do not produce large changes in prediction sensitivity.

This section describes \emph{smooth prediction sensitivity}, an extension of prediction sensitivity that is \emph{smooth by design}.

\paragraph{Techniques for smoothing gradients.}
To achieve a smooth estimation of prediction sensitivity, we adapt methods from interpretability for deep learning. Our approach is most similar to \smoothgrad~\cite{smilkov2017smoothgrad}, which uses the gradient of a prediction to determine which features were most important in making that prediction. \smoothgrad (and related approaches for interpretability~\cite{morch1995visualization, baehrens2010explain, simonyan2013deep, sundararajan2017axiomatic, hooker2019benchmark}) are most often applied in image classification tasks, where each feature corresponds to a pixel, to build visualizations that show which regions of the image were most important for classifying it. In the image classification setting, non-smooth gradients translate into noisy visualizations; \smoothgrad produces visualizations with less noise by smoothing the gradient.

The main idea of \smoothgrad is to sample many \emph{similar} examples to the one under consideration, by perturbing the example slightly with Gaussian noise. Then, \smoothgrad calculates the gradient of the prediction on each of the perturbed examples and returns the average. Formally, \smoothgrad calculates the following, where $n$ is the number of samples to take:
\[\text{\smoothgrad}(x) = \frac{1}{n} \sum_{i = 1}^n \nabla_x(\mathcal{F}(\theta, x + \mathcal{N}(0, \sigma^2))) \]
\noindent Smilkov et al.~\cite{smilkov2017smoothgrad} conduct an experimental evaluation in the image classification setting that shows (1) that gradients are sometimes \emph{not} smooth, suggesting that smoothing might be helpful, and (2) that the \smoothgrad approach produces visualizations with less noise in these cases. Later empirical analysis~\cite{hooker2019benchmark} reinforces the benefits of this approach.

\paragraph{Smoothing prediction sensitivity.}
We apply the \smoothgrad approach to prediction sensitivity to calculate \emph{smooth prediction sensitivity}, a measure of individual fairness that is robust in the face of rapidly fluctuating gradients. Our setting has a major difference from the image classification setting of \smoothgrad: to measure fairness, we would like a \emph{worst-case} analysis, rather than an average-case one (i.e. we would like to measure the \emph{worst} prediction sensitivity in the neighborhood of the example under consideration, rather than the average). We therefore define smooth prediction sensitivity as follows:
\[ \text{\sc SmoothPredictionSensitivity}(x) = \max_{i \in {1 \dots n}} \Big\lvert \frac{\partial}{\partial a} \mathcal{F}(\theta, x + \mathcal{N}(0, \sigma^2)) \Big\rvert \]
\noindent Our definition of smooth prediction sensitivity captures the conceptual goal of measuring the worst-case impact of the protected attribute's value in the neighborhood around a specific example.

\paragraph{Limitations \& future work.}
Smooth prediction sensitivity has several important limitations, which we outline here. First, sampling $n$ points in a (potentially) very high-dimensional space may easily miss regions in the example's neighborhood with higher prediction sensitivity, causing our definition to underestimate the true worst-case sensitivity. Second, the variance of the noise used in sampling ($\sigma$) defines the size of the neighborhood for our search, and the best value for $\sigma$ is likely to depend on the model architecture and data. While our preliminary experimental results (discussed next) suggest that reasonable ``default'' values for $n$ and $\sigma$ may work well across models and datasets, further study is needed to understand their role completely.

In addition, smooth prediction sensitivity shares an important limitation with the original formulation of prediction sensitivity: it measures \emph{only} the prediction's reliance on the protected attribute. Models may make unfair predictions \emph{without} consulting the protected attribute (e.g. if other attributes are correlated with the protected attribute), and our approach would not be able to distinguish these predictions from entirely fair ones. Moreover, our approach is limited to datasets with an explicit protected attribute, and is therefore not applicable in settings where the protected attribute is not explicit (e.g. natural language or image domains). These two limitations---correlated features and implicit protected attributes---represent important for future work.

\section{Evaluation}

Our empirical evaluation investigates the following research questions:
\begin{enumerate}[topsep=-1mm,leftmargin=6mm]
\itemsep0.0mm
\item Does prediction sensitivity accurately distinguish between fair
  and unfair predictions?
\item Does smooth prediction sensitivity demonstrate an increased
  ability to make this distinction?
\item Can prediction sensitivity detect unfair predictions, even when
  they are made by models trained with fairness mitigations?
\end{enumerate}

Our results provide positive evidence for each of these three
questions. The results show a difference between the distributions of
prediction sensitivities on fair and unfair predictions, and that
smoothing preserves and sometimes improves the difference. Finally, we
show that predictions with high prediction sensitivity exist in
practice, even for models trained with a fairness mitigation.

\paragraph{Datasets.}
Our evaluation considers two commonly-used datasets in the AI fairness
literature: the Adult~\cite{adult} and COMPAS~\cite{compas}
datasets. Statistics of the two datasets are shown in
Figure~\ref{fig:datasets}. Both involve classification tasks: the
Adult dataset's task is predicting whether an individual's income is
greater than or less than \$50,000, and the COMPAS dataset's task is
predicting recidivism risk. Both datasets are known to contain
embedded bias: models trained on the Adult dataset tend to predict
that white males have the highest chance of having high income, and
models trained on the COMPAS dataset tend to predict that Black
suspects are most likely to re-offend.

\begin{figure}
  \centering
  \begin{tabular}{|l| l l| l l | l l |}
    \hline
    \textbf{Dataset}
    & \multicolumn{2}{c|}{\textbf{Dataset Size}}
    & \multicolumn{2}{c|}{\textbf{Linear Model}}
    & \multicolumn{2}{c|}{\textbf{CNN Model}} \\
    & Train & Test
    & Match & !Match
    & Match & !Match \\
    \hline
    Adult~\cite{adult}
    & 24752 
    & 6188  
    & 5173  
    & 1015  
    & 4908  
    & 1280  
    \\
    COMPAS~\cite{compas}
    & 4937  
    & 1235  
    & 854   
    & 381   
    & 813   
    & 422   
    \\
    \hline
  \end{tabular}
  \caption{Dataset statistics and ``ground truth'' for fairness. \emph{Match} and \emph{!Match} denote the number of testing examples where the ``fair'' and ``unfair'' models agree and disagree, respectively.}
  \label{fig:datasets}
\end{figure}

\paragraph{Model architectures.}
We used two model architectures in our experiments: a linear network and a convolutional neural network(cnn). 
The linear network architecture comprises 3 hidden linear layers with 32 neurons each ReLU, followed by a linear output layer with a single neuron. The ReLU activation function is used on the hidden layers while the sigmoid activation function is applied on the output layer. To avoid overfitting a Dropout of 0.2 is used on each hidden layer.  

The cnn network architecture can be described using the shorthand notation C256-C128-C64-C32-C16-C1 where C refers to a 1D-Convolutional layer and the number indicates the number of filters. Each layer is followed by the ReLU activation except the last/output layer that uses a sigmoid activation function.

The loss function used for both architectures is the binary cross entropy loss and Adam is chosen as the optimizer with a learning rate of 0.001.

\paragraph{Fair and unfair models.}
We trained two models for each of our experiments: an \emph{unfair}
model, trained without any fairness mitigations, and a \emph{fair}
model, trained using a fairness mitigation approach that was proposed by Louppe et al.~\cite{louppe2017learning}. Their approach,  which was inspired by Generative Adversarial Networks (GANs), leverages adversarial networks to enforce a so called pivotal property on a predictive model. In this method the authors propose a hyperparameter that controls the trade-off between accuracy and robustness.  This pivotal property ensures that the outcome distribution of the model is independent of the protected attributes, thereby producing fair predictions. 

For each model architecture in our experiments, we created a corresponding adversarial network. The adversarial network architectures are identical to the model architectures at all levels except in the output layer. In the adversarial network, the number of neurons in the output layer is equivalent to the number of protected attributes in the dataset. Additionally, all dropout layers were removed. 

In this work, we use 4 metrics of group fairness: disparate impact~\cite{feldman2015certifying}, statistical parity~\cite{10.1145/2090236.2090255}, equality of opportunity~\cite{hardt2016equality} and average odds difference.

\paragraph{Ground truth for ``fair'' and ``unfair'' predictions.}
No ground truth datasets exist for distinguishing between fair and
unfair predictions at the individual level. To construct a proxy for
this data, we used the ``fair'' and ``unfair'' models described
above. We used both models to make predictions on the test set for
each of our datasets, and then compared those predictions. We labeled
a prediction ``fair'' if the two models agreed on it; we labeled a
prediction ``unfair'' if the two models
disagreed. Figure~\ref{fig:datasets} summarizes the number of ``fair''
and ``unfair'' predictions for each model architecture and dataset
combination.

It is important to note that this approach is only a crude
approximation of a notion of fairness defined \emph{in terms of the
  mitigation used}. We use this approach for evaluation because no
ground truth data for individual predictions is available. We describe
``fair'' and ``unfair'' predictions using quotation marks, to
emphasize that \textbf{these categories do \emph{not} reference actual
  ground truth descriptions of fairness}.

\newcommand{\gwidth}{.45\textwidth}
\begin{figure}
  \begin{tabular}{c c c}
    & \textbf{Adult}  & \textbf{COMPAS}\\
    \rotatebox[origin=l]{90}{\hspace*{15mm}\textbf{Linear Model}}
    & \includegraphics[width=\gwidth]{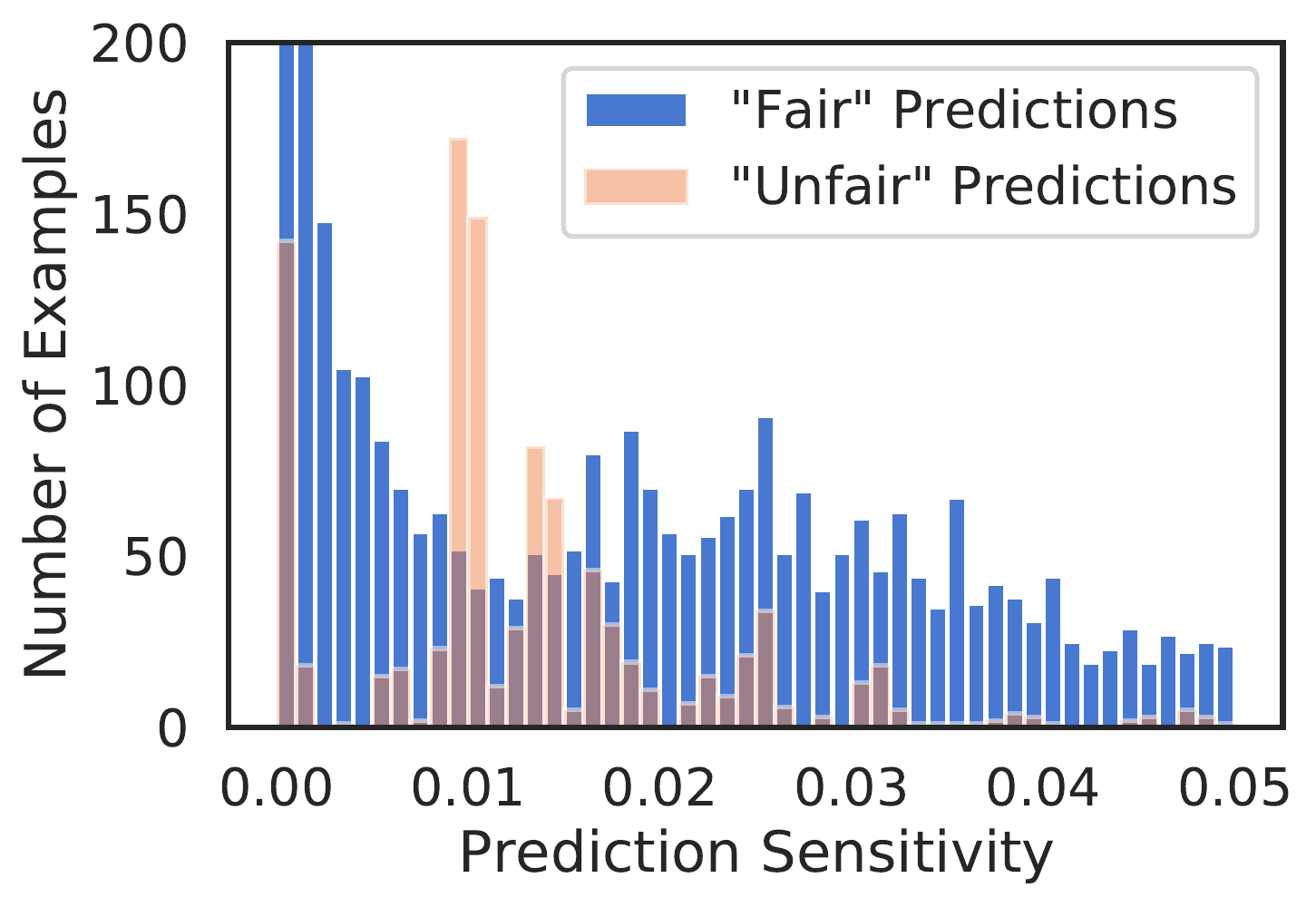}
    & \includegraphics[width=\gwidth]{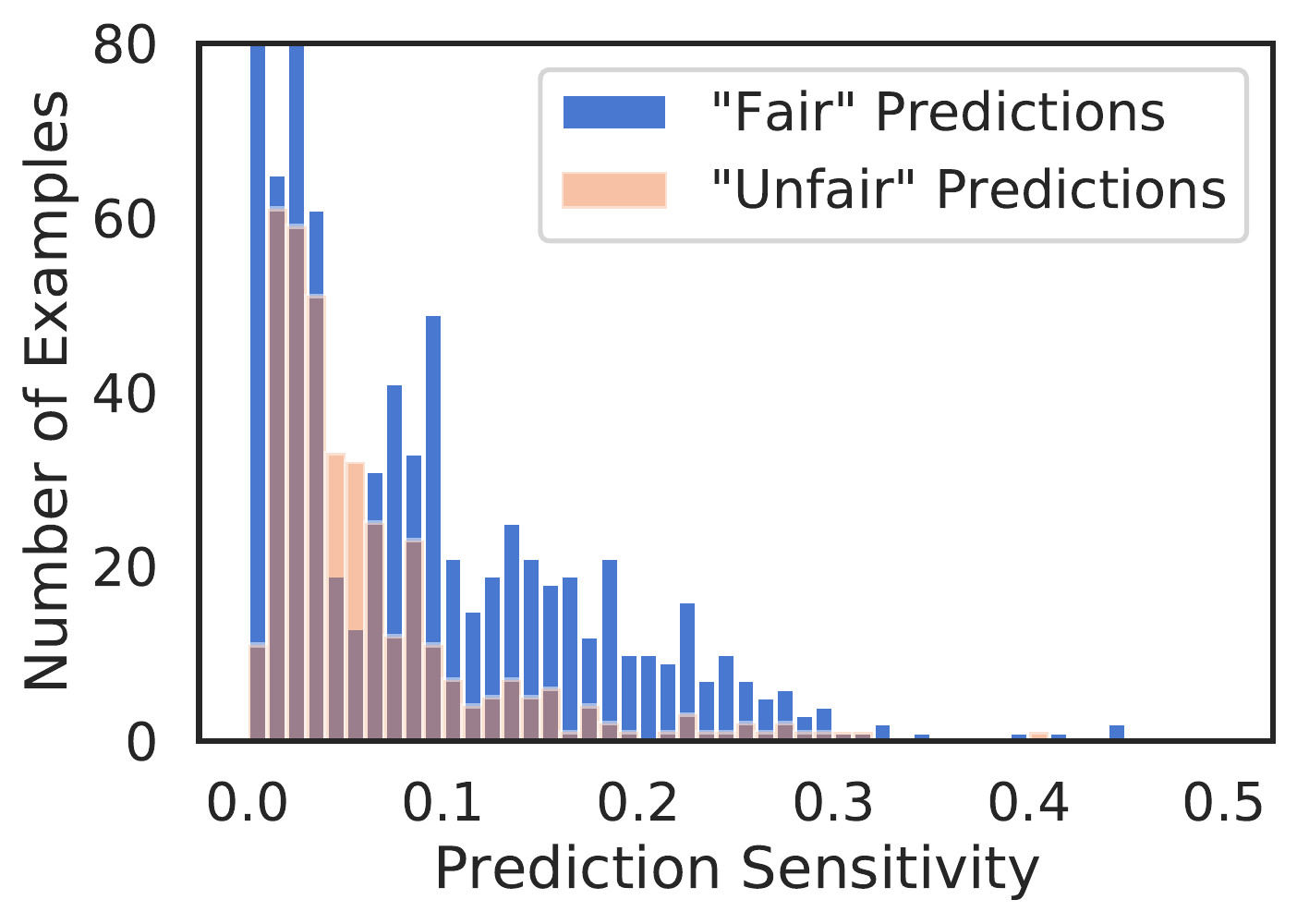} \\
    \rotatebox[origin=l]{90}{\hspace*{15mm}\textbf{CNN Model}}
    & \includegraphics[width=\gwidth]{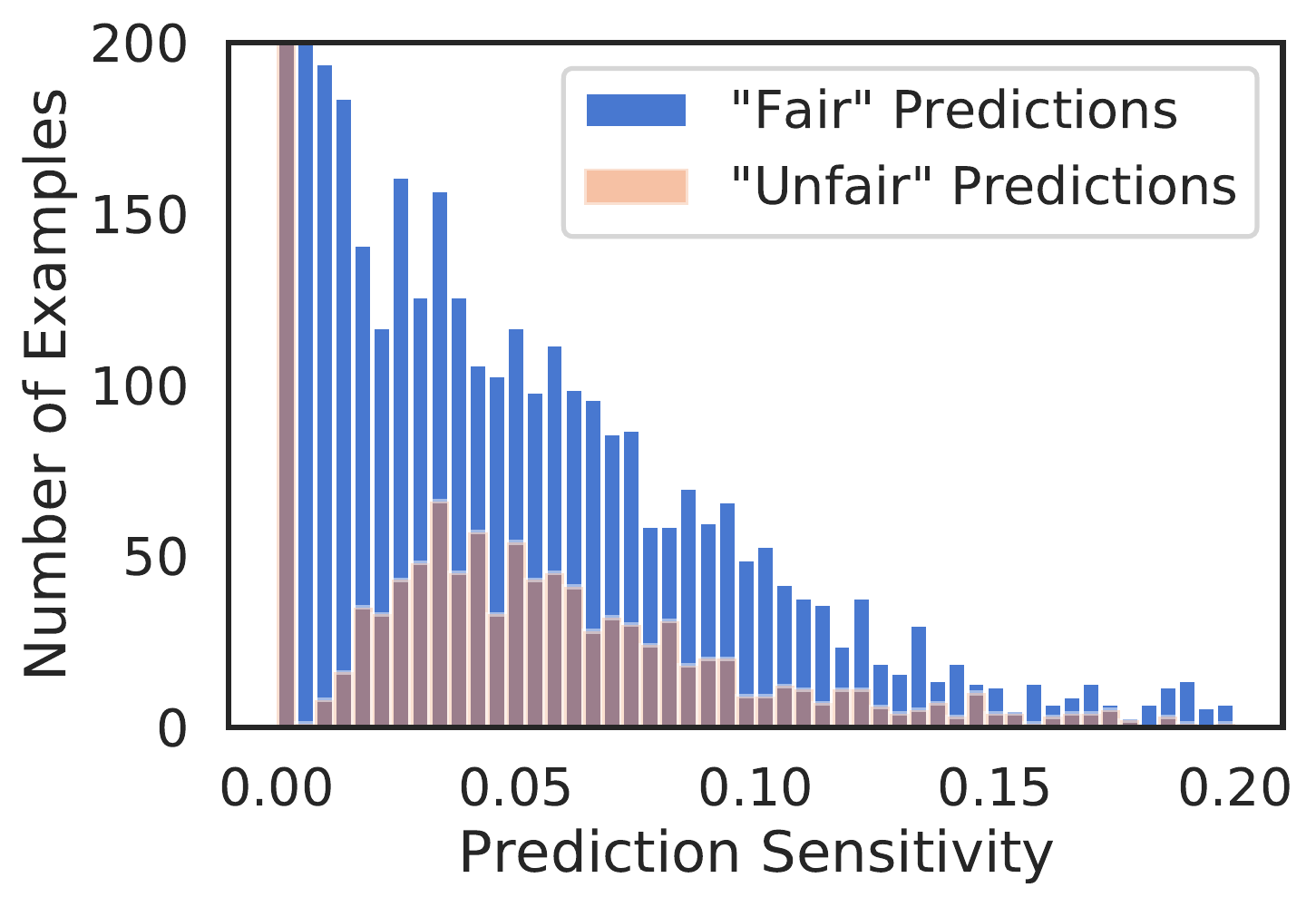}
    & \includegraphics[width=\gwidth]{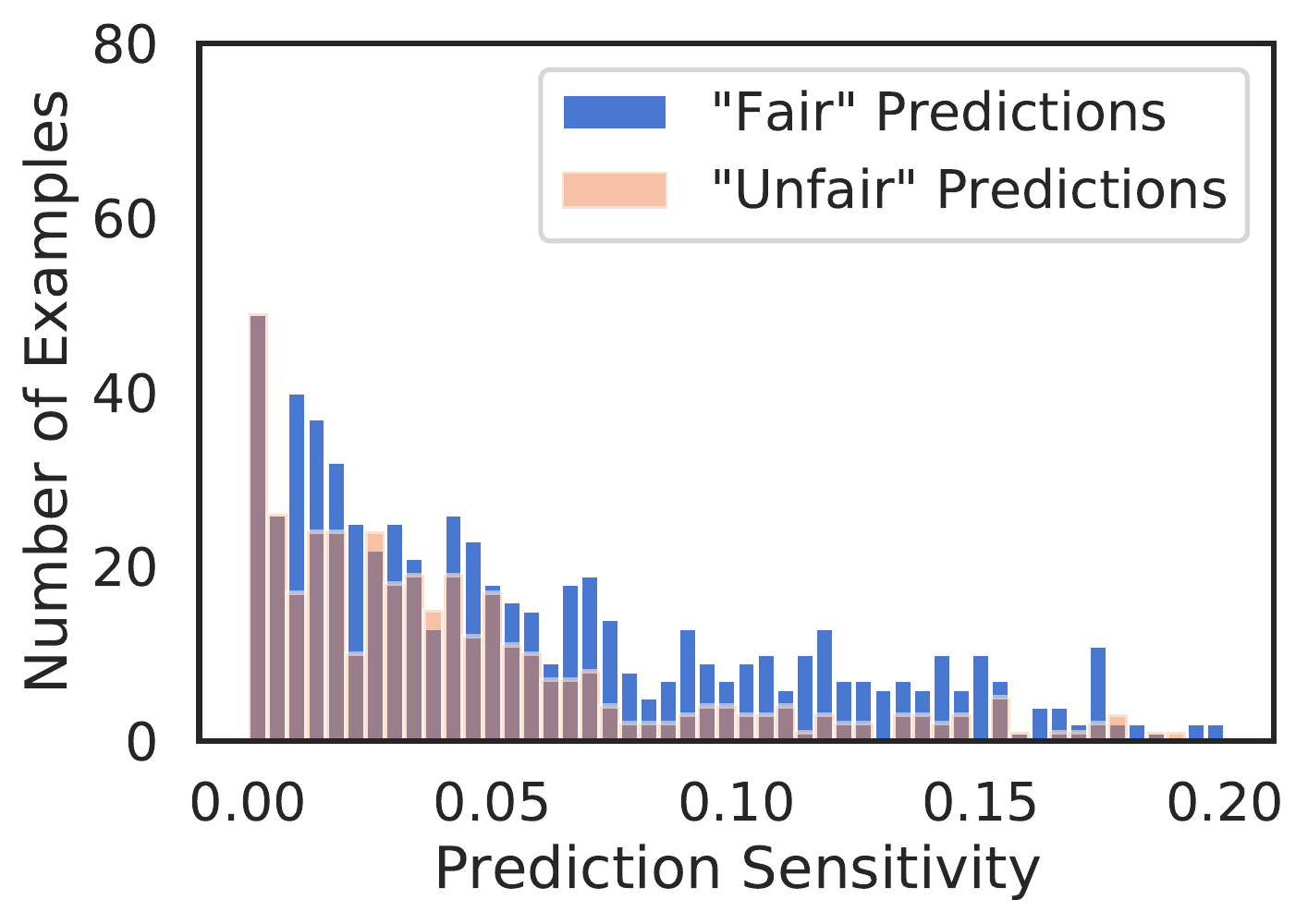} \\
  \end{tabular}

  \caption{\label{fig:results_unsmoothed_unfair}\textbf{Un-smooth prediction sensitivity} for ``fair'' and ``unfair'' predictions, \textbf{no mitigation}.}
\end{figure}

\begin{figure}
  \begin{tabular}{c c c}
    & \textbf{Adult}  & \textbf{COMPAS}\\
    \rotatebox[origin=l]{90}{\hspace*{15mm}\textbf{Linear Model}}
    & \includegraphics[width=\gwidth]{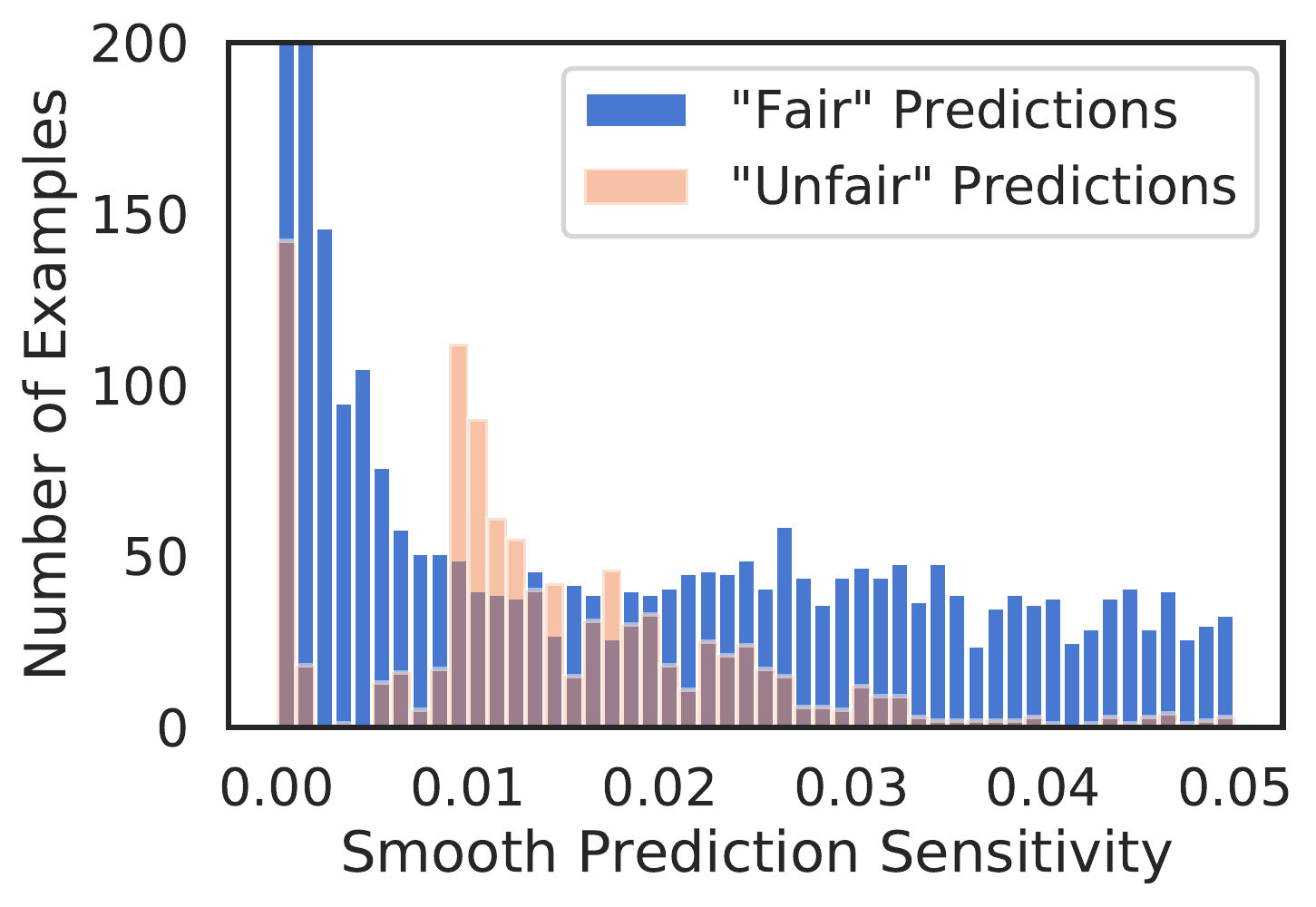} & \includegraphics[width=\gwidth]{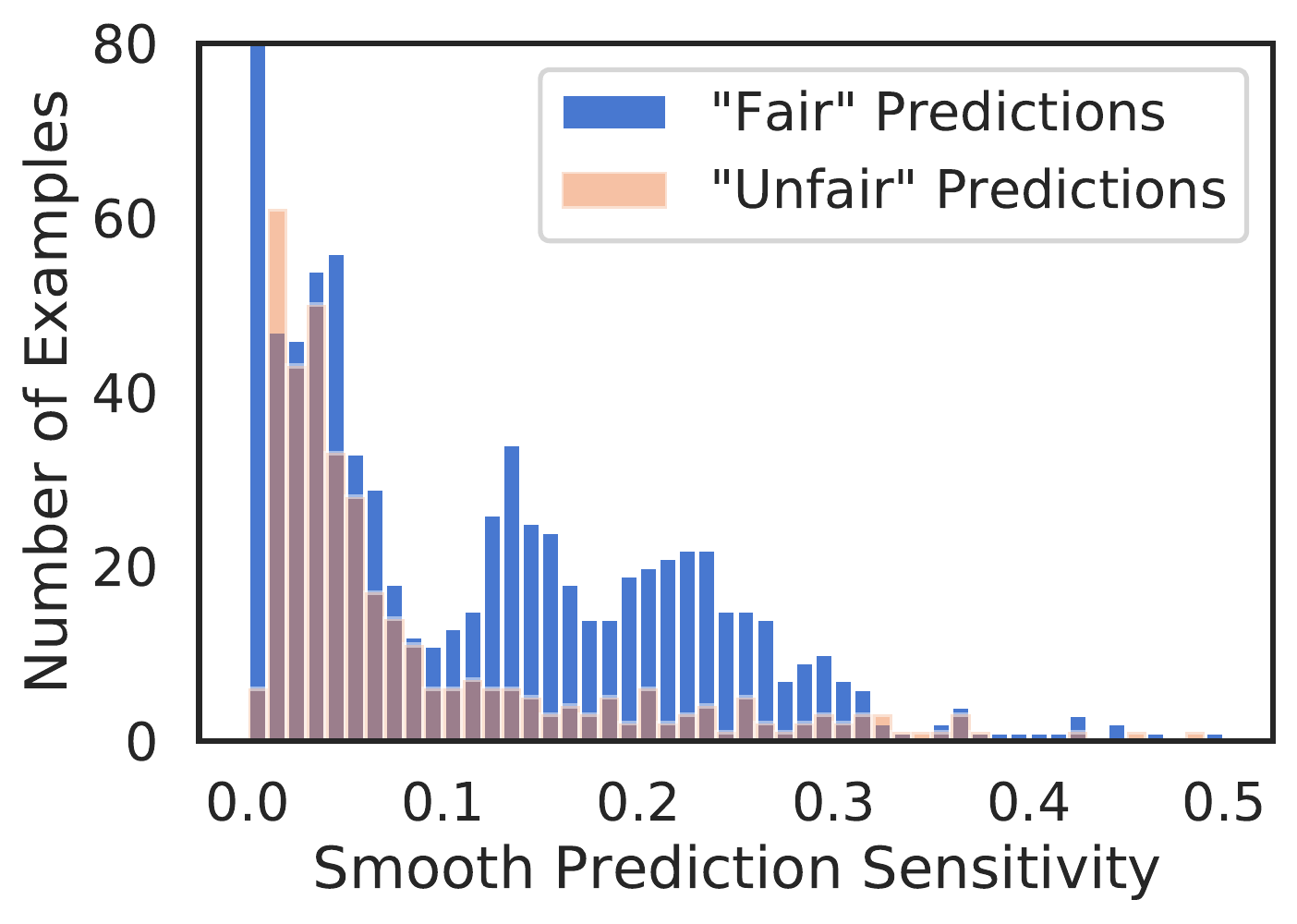} \\
    \rotatebox[origin=l]{90}{\hspace*{15mm}\textbf{CNN Model}}
    & \includegraphics[width=\gwidth]{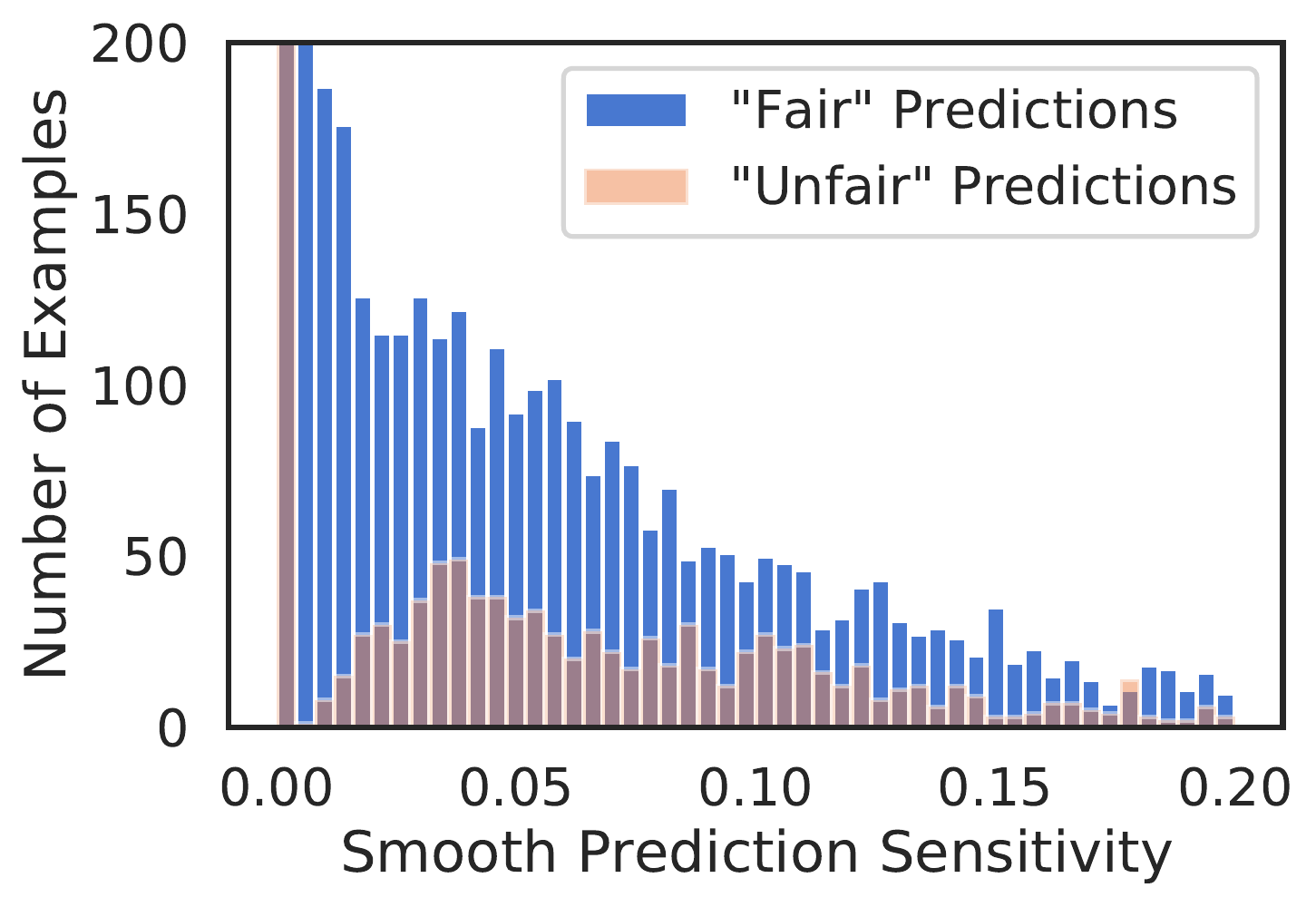} &  \includegraphics[width=\gwidth]{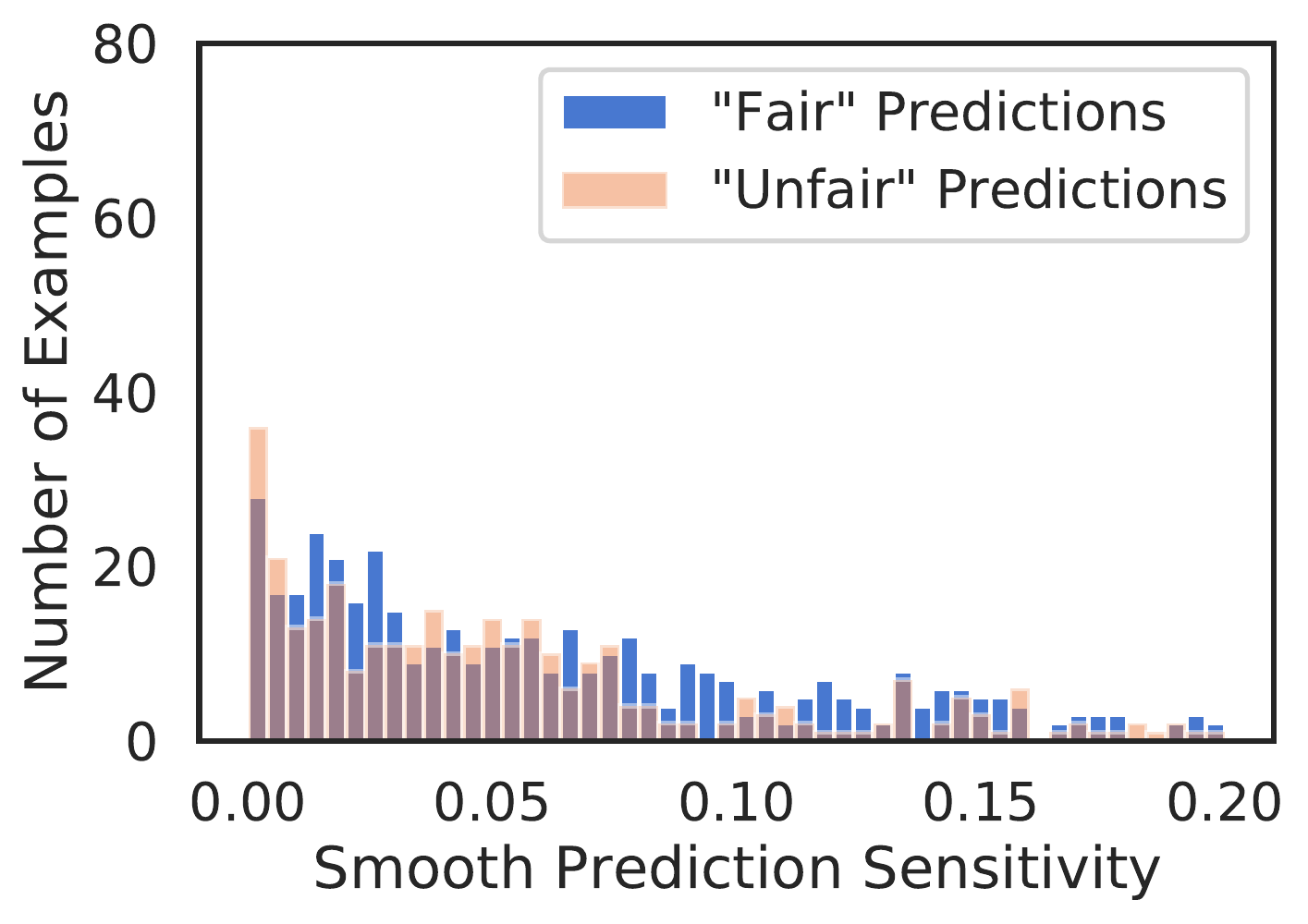} \\
  \end{tabular}

  \caption{\label{fig:results_smoothed_unfair}\textbf{Smooth prediction sensitivity} for ``fair'' and ``unfair'' predictions, \textbf{no mitigation}.}
\end{figure}



\begin{figure}
  \begin{tabular}{c c c}
    & \textbf{Adult}  & \textbf{COMPAS}\\
    \rotatebox[origin=l]{90}{\hspace*{15mm}\textbf{Linear Model}}
    & \includegraphics[width=\gwidth]{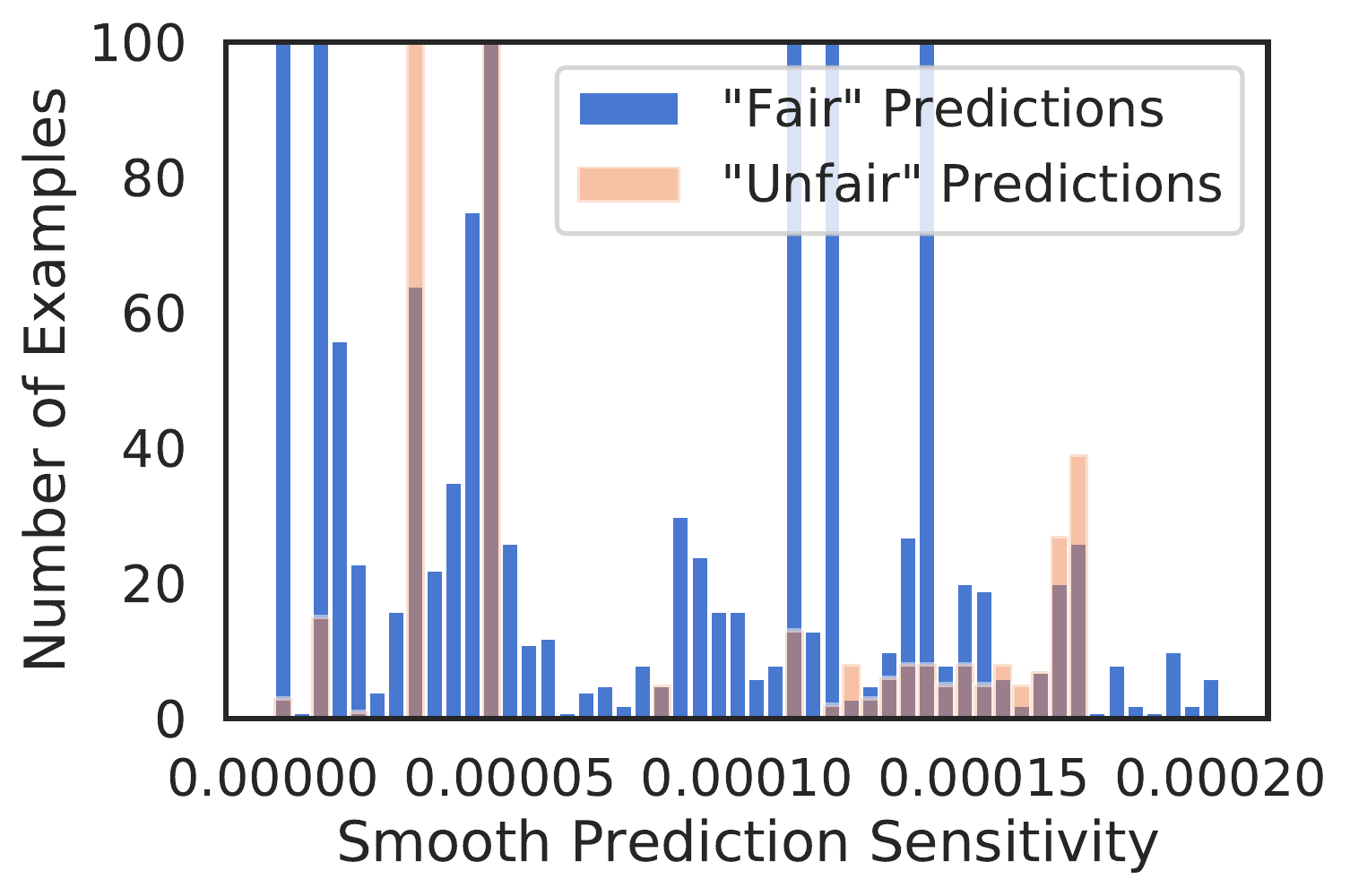}
                      & \includegraphics[width=\gwidth]{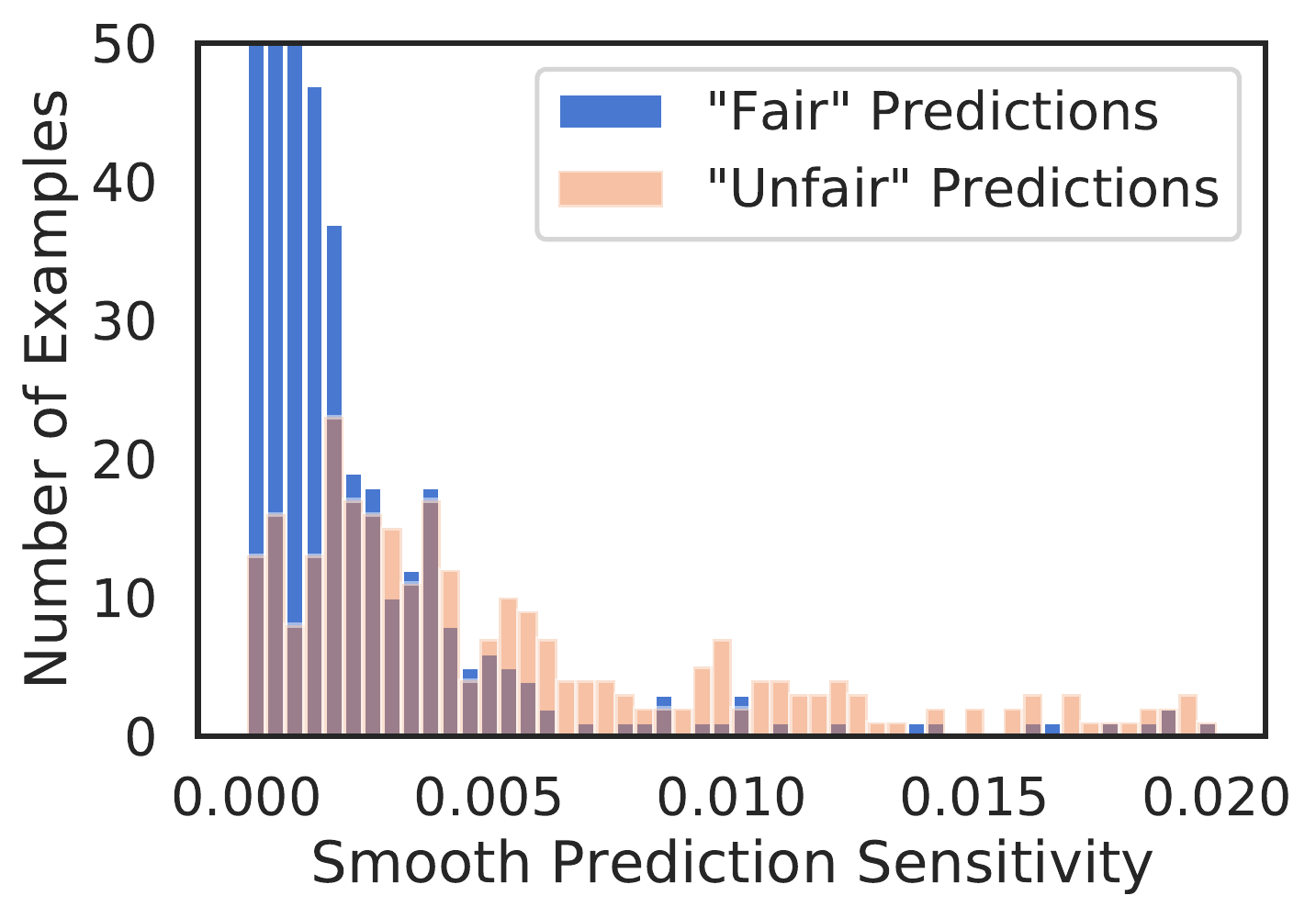} \\
    \rotatebox[origin=l]{90}{\hspace*{15mm}\textbf{CNN Model}}
    & \includegraphics[width=\gwidth]{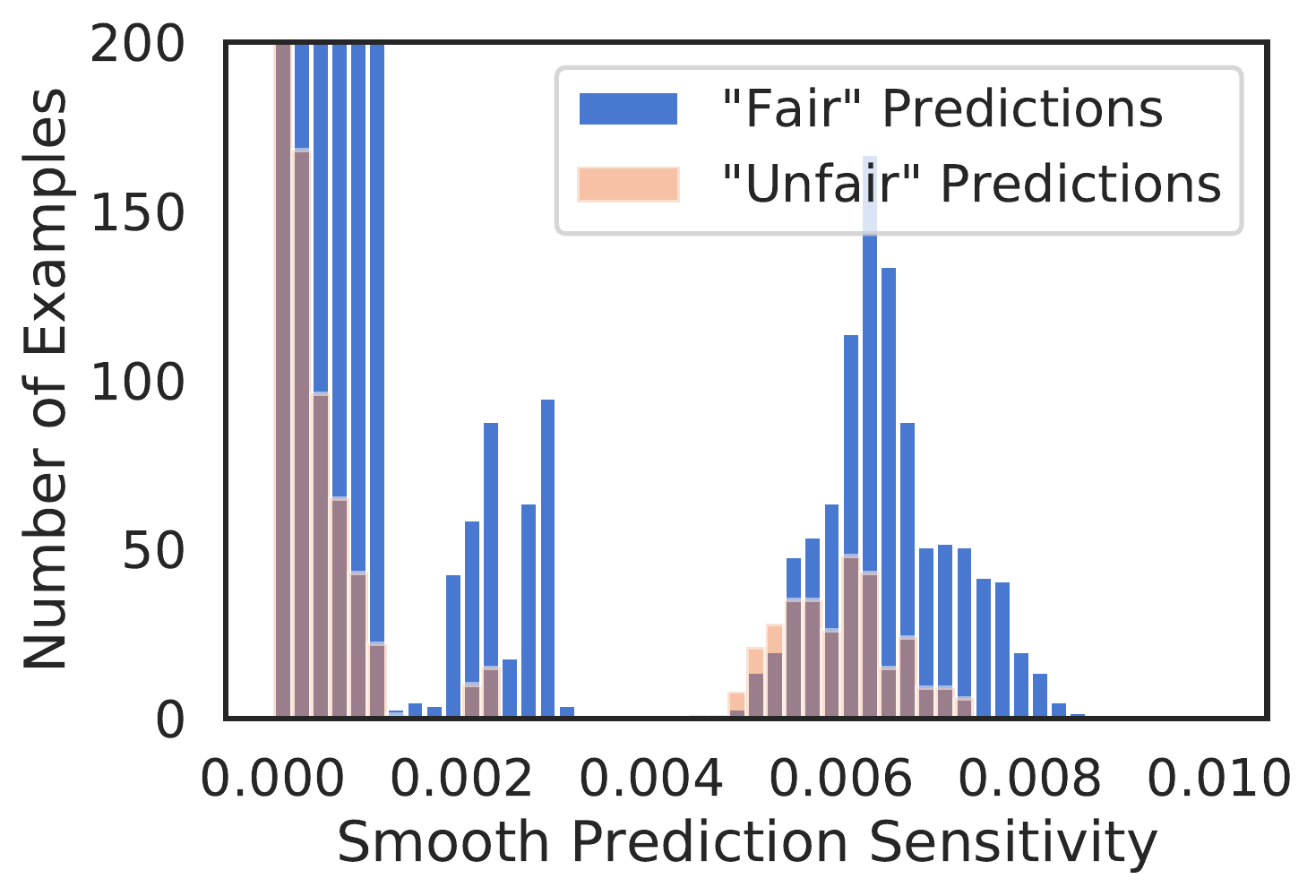} & \includegraphics[width=\gwidth]{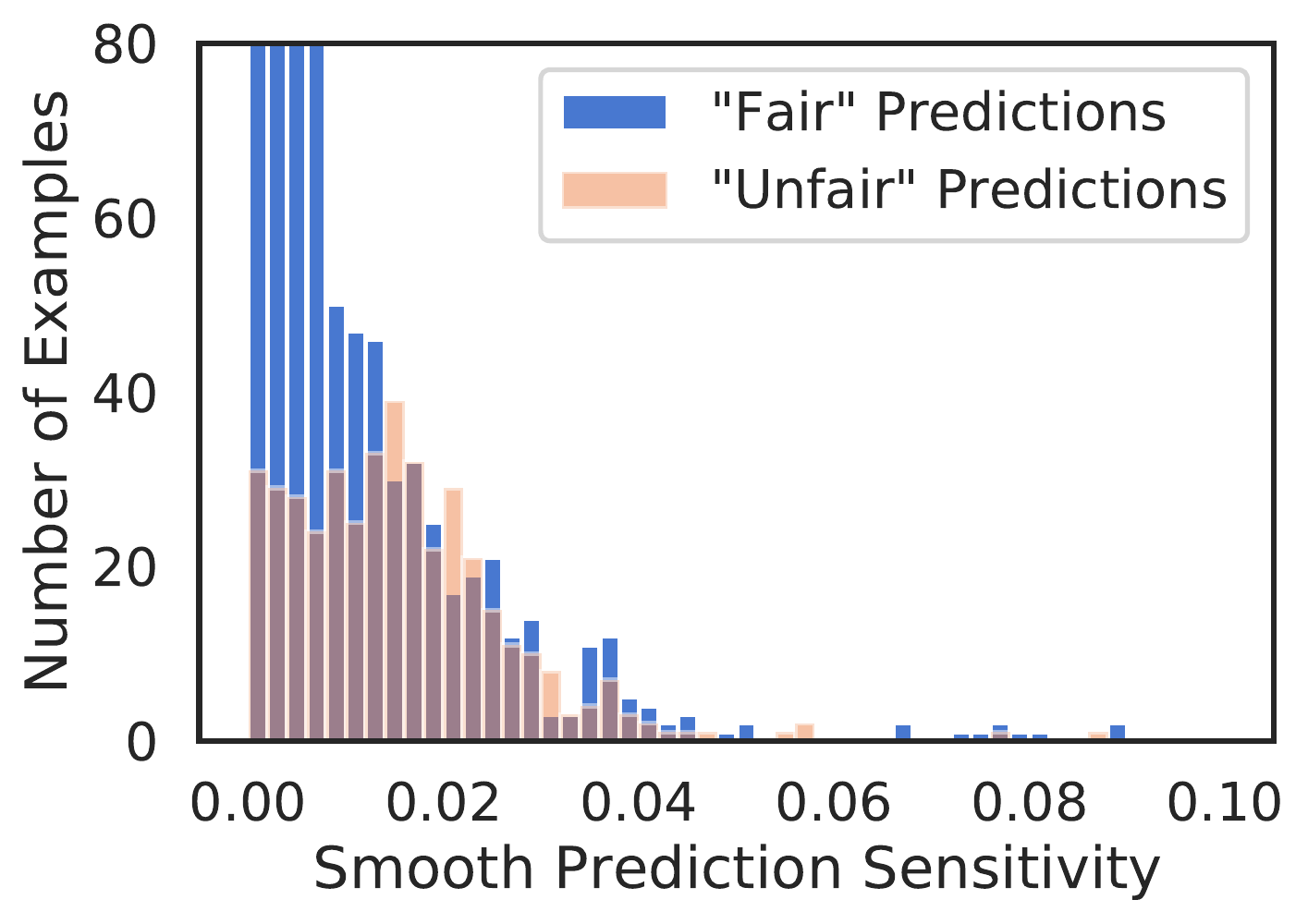} \\
  \end{tabular}

  \caption{\label{fig:results_smoothed_fair}\textbf{smooth prediction sensitivity} for ``fair'' and ``unfair'' predictions, \textbf{with mitigation}.}
\end{figure}

\begin{figure}
  \begin{tabular}{c c c}
    & \textbf{Adult}
    & \textbf{COMPAS}\\
    \rotatebox[origin=l]{90}{\hspace*{15mm}\textbf{Linear Model}}
    & \includegraphics[width=\gwidth]{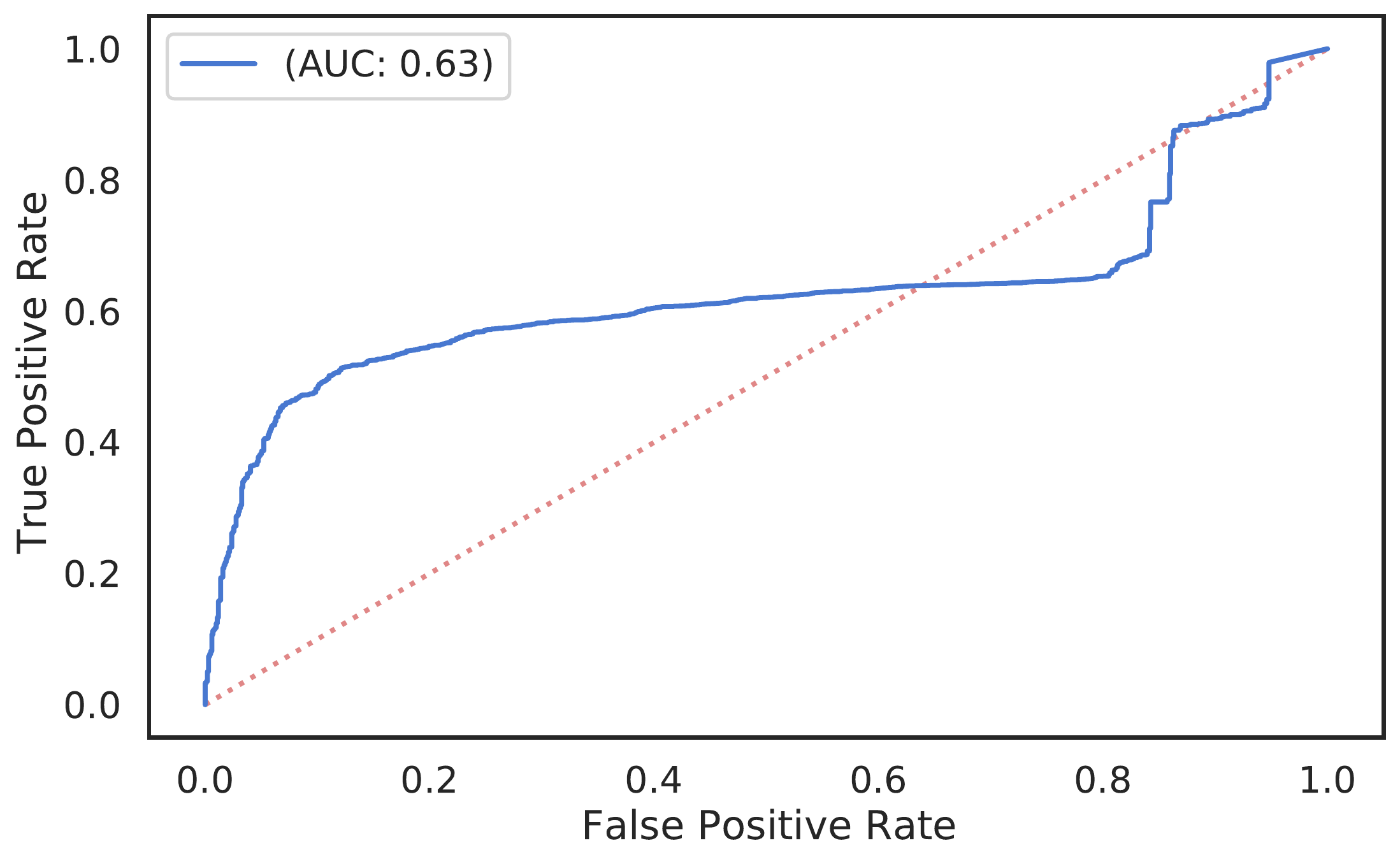}
    & \includegraphics[width=\gwidth]{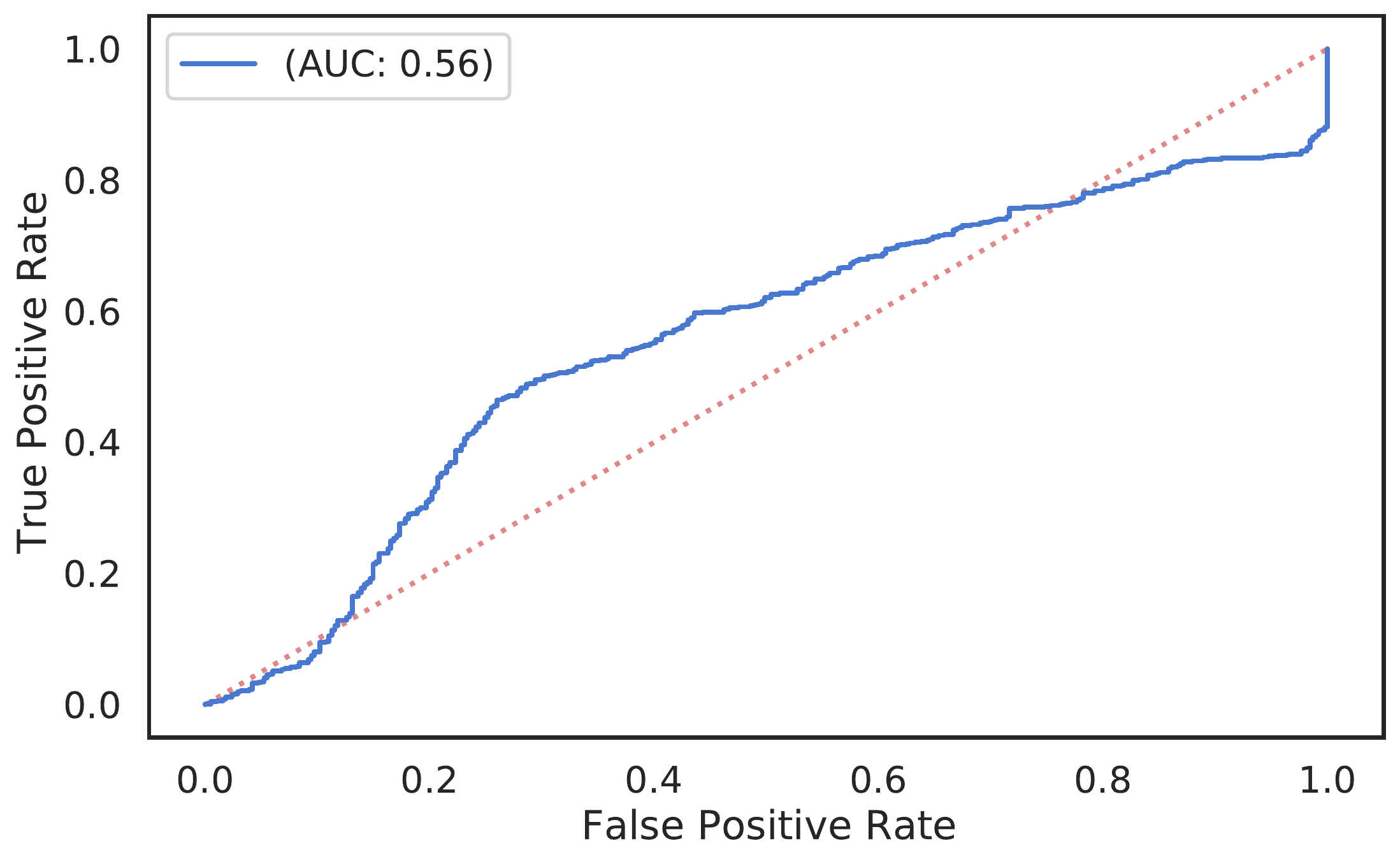} \\
    \rotatebox[origin=l]{90}{\hspace*{15mm}\textbf{CNN Model}}
    & \includegraphics[width=\gwidth]{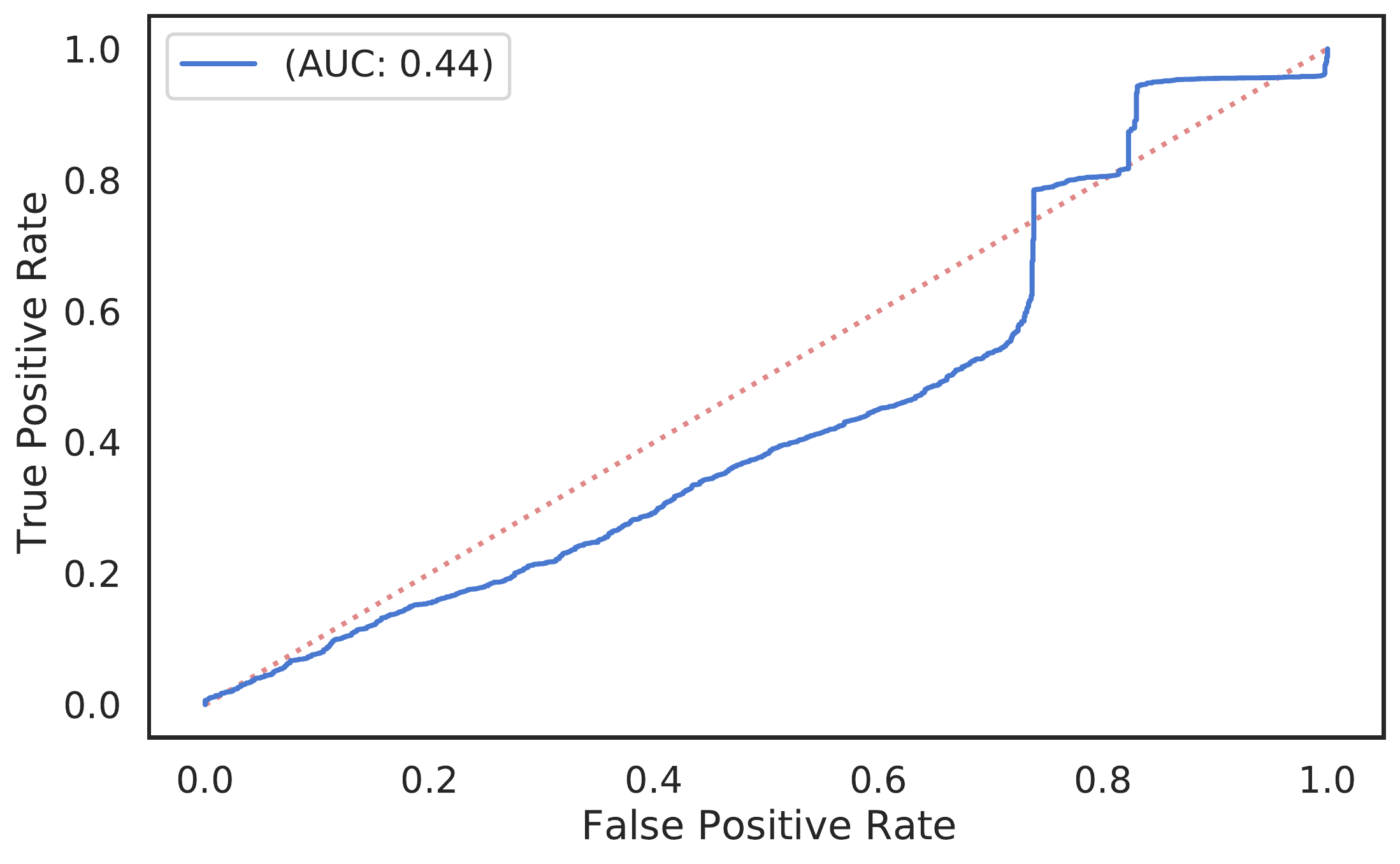}
    & \includegraphics[width=\gwidth]{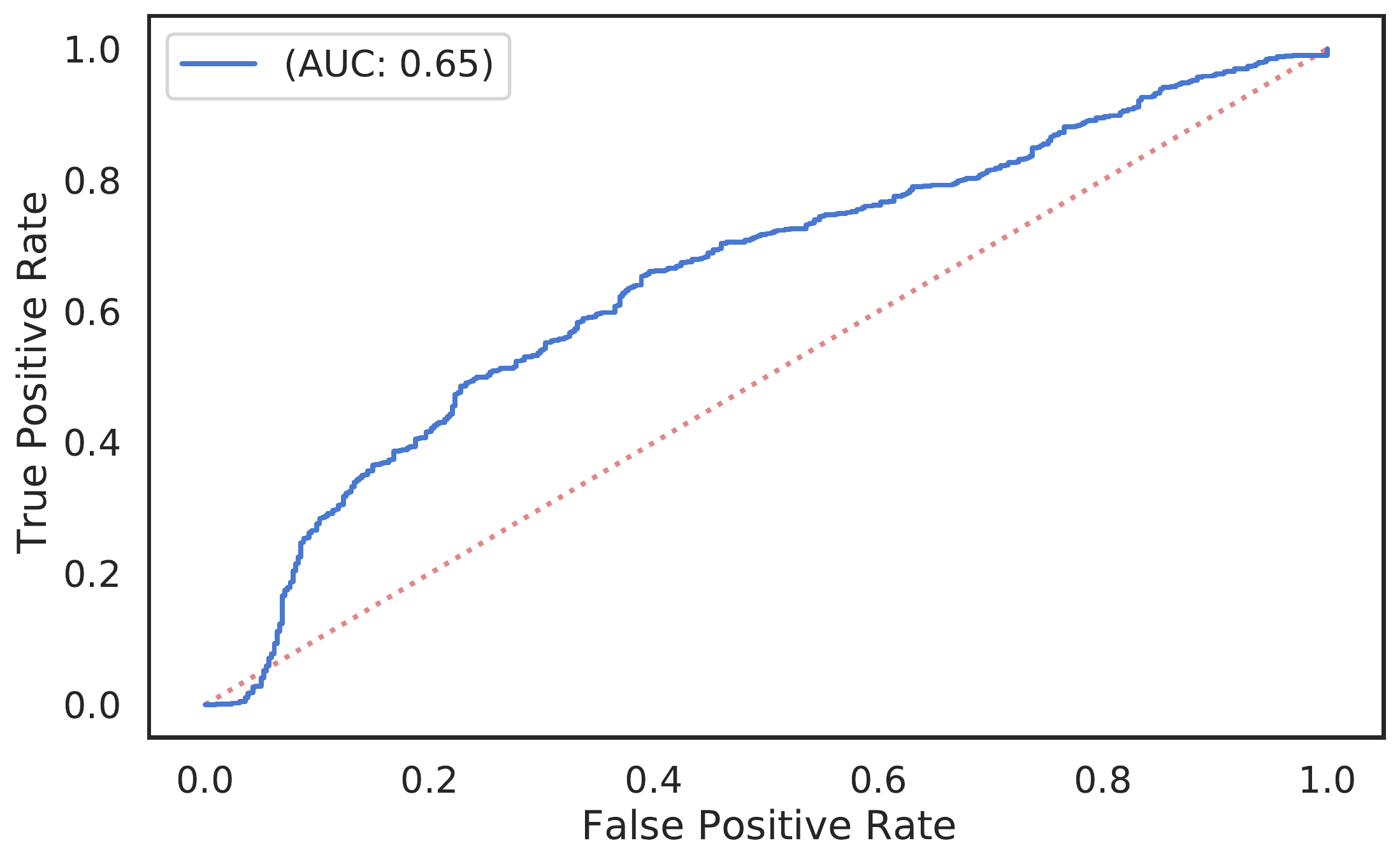} \\
  \end{tabular}

  \caption{\label{fig:roc_curves}ROC curves: predicting individual fairness using smooth prediction sensitivity.}
\end{figure}

\paragraph{Results.}
Figure~\ref{fig:results_unsmoothed_unfair} presents our experimental
results using prediction sensitivity on the \emph{unfair} model
(trained without mitigation). The blue and orange bars indicate the
distribution of prediction sensitivities for the ``fair'' and
``unfair'' predictions made by the model, respectively. The results
are mixed: on the Adult dataset, for both model architectures, the
distribution of prediction sensitivities for ``fair'' and ``unfair''
predictions have different shapes; on the COMPAS dataset, the two
distributions look fairly similar. For both datasets, the average
prediction sensitivity for ``unfair'' predictions is higher than for
``fair'' predictions. The two results suggest that prediction
sensitivity may indeed be a good match for existing formal fairness
metrics.

Figure~\ref{fig:results_smoothed_unfair} shows the effect of smoothing
on prediction sensitivity. The results are similar to the case with no
smoothing, suggesting that prediction sensitivity is \emph{already}
fairly smooth for the model architectures we used. Smoothing
\emph{does} appear to smooth out the distributions of sensitivities in
some cases (e.g. the linear model on the Adult dataset), suggesting
that it might be useful in some cases.

Figure~\ref{fig:results_smoothed_fair} shows similar results, but for
the models trained \emph{with the fairness mitigation}. The results
show that the vast majority of examples now have low prediction
sensitivity, providing further evidence that prediction sensitivity
aligns with existing group fairness metrics. However, for the CNN
model on the Adult dataset, the results also show a cluster of
predictions with \emph{high} prediction sensitivity. This cluster
suggests that the model may be making unfair predictions, \emph{even
  after the mitigation has been applied}, and that prediction
sensitivity may be effective in detecting this condition. This is a
particularly difficult point to evaluate, due to the limitations of
the ground truth labels, and it is one of our targets for future work.

Figure~\ref{fig:roc_curves} shows how effective smooth prediction
sensitivity is as a \emph{classifier} for ``fair'' and ``unfair''
predictions. The results suggest that prediction sensitivity provides
some capability to distinguish between ``fair'' and ``unfair''
predictions, but that its accuracy for this task is fairly
modest. Note that \textbf{prediction sensitivity is not intended to be
  applied this way}, and using it (or any formal fairness measure) in
an automated decision-making process could potentially lead to
additional harm to society.

Our empirical results provide preliminary evidence that prediction
sensitivity is well-aligned with common notions of group privacy; that
smooth prediction sensitivity maintains the properties of
un-smooth prediction sensitivity, and can help smooth out its
values; and that prediction sensitivity may even be capable of
detecting when models trained with fairness mitigations make blatantly
unfair predictions to individuals.

\section{Conclusion}

We have presented smooth prediction sensitivity, which builds on
prediction sensitivity to provide a measure of individual fairness to
enable auditability for fairness of predictions made by deep learning
systems. Our empirical results provide preliminary evidence that
smooth prediction sensitivity aligns well with existing metrics for
fairness---suggesting that it may provide a useful measure of fairness
in practice---but also show that further study is needed to better
understand how to apply the measure. Our results also suggest that
models trained with fairness mitigation techniques may nevertheless
make some unfair predictions, and that prediction sensitivity may be
useful in detecting this phenomenon.




\section*{Broader Impact}

In 2020, members of the American Mathematical Society (AMS) signed a letter demanding that ``any algorithm with potential high impact face a public audit.'' Our work represents a step in this direction, and we hope that it will lead to an increased ability to audit the decisions made by AI-based systems for fairness.

However, the letter also cites the potential for algorithms that rely on solely mathematical assessments for fairness to be used to create a ``scientific veneer for racism.'' We acknowledge that the broader impact within which our work can be contextualized involves the possibility that formally defined measures can be weaponized against communities they purport to protect and used to defend unfair decisions, aggregating an abuse of power. Improperly applied in an automated system, our approach could make the situation \emph{worse} by acting as a scientific veneer. We believe that our work can be helpful for \emph{humans} to perform audits of AI-based systems, but caution that it is not suitable for use in automated systems.

Moreover, the conclusion that mathematical definitions of fairness alone can determine fairness is a misguided one, and we stand in agreement with communities such as the AMS that have identified issues of tech washing that make algorithmic prediction technology---particularly those that have been proven to be used in support of structural racism and oppression---highly problematic.

With our research, we present a contribution to the field of fairness that speaks to the need for auditing systems that can assist in overseeing and providing feedback in a way that is interpretable and upholds the legal due process for democracy. This feedback  affects the legal and social implications of deep learning models that are used in human processes.

\section*{Acknowledgments}

We thank David Darais and Kristin Mills for their contributions to the
development of this work, and the Algorithmic Fairness through the
Lens of Causality and Interpretability (AFCI) reviewers for their
suggestions for improvement. This research was supported in part by an
Amazon Research Award.


\bibliographystyle{plain}
\bibliography{refs, fairness}

\end{document}